\documentclass[10pt, a4paper]{article}

\usepackage[table]{xcolor}
\usepackage{todonotes}
\usepackage{amsmath}
\usepackage{comment}
\usepackage{float}
\usetikzlibrary{chains,shapes.geometric, positioning,fit,arrows.meta,backgrounds,arrows,calc}

\def\SB{{\sc MEDIA 2022}}
\def\HA{{\sc HumanE-AI-Net}} 

\usepackage[final]{lrec-coling2024} 

\title{New Semantic Task for the French Spoken Language Understanding MEDIA Benchmark}
\name{Nadège Alavoine$^1$, Gaëlle Laperrière$^2$, Christophe Servan$^{1,2}$, \\ \large \bf Sahar Ghannay$^{1}$ and Sophie Rosset$^1$}

\address{$^1$Université Paris-Saclay, CNRS, LISN, $^2$Avignon Université, LIA, $^3$QWANT\\
         \{firstname.lastname\}@lisn.upsaclay.fr\\}
         
\abstract{Intent classification and slot-filling are essential tasks of Spoken Language Understanding (SLU).
In most SLU systems, those tasks are realized by independent modules.
For about fifteen years, models achieving both of them jointly and exploiting their mutual enhancement have been proposed.
A multilingual module using a joint model was envisioned to create a touristic dialogue system for a European project, \HA{}.
A combination of multiple datasets, including the MEDIA dataset, was suggested for training this joint model.
The MEDIA SLU dataset is a French dataset distributed since 2005 by ELRA, mainly used by the French research community and free for academic research since 2020. Unfortunately, it is annotated only in slots but not intents.
An enhanced version of MEDIA annotated with intents has been built to extend its use to more tasks and use cases.
This paper presents the semi-automatic methodology used to obtain this enhanced version. 
In addition, we present the first results of SLU experiments on this enhanced dataset using joint models for intent classification and slot-filling.
 \\ \newline  }

\begin{document}

\maketitleabstract

\section{Introduction}


The Spoken Language Understanding (SLU) module is a crucial component of a spoken language dialogue system. 
It semantically analyzes user queries and identifies speech or text spans that mention semantic information. 
SLU tasks can fall into three sub-tasks: domain classification, intent classification, and slot-filling \cite{tur2011}. 
In this study, we are interested in intent classification and slot-filling tasks. The latter task can also be considered as a concept detection task \cite{BonneauMaynard2006}.

Most dialogue systems handle those tasks separately by developing independent modules inserted in a pipeline \cite{HakkaniTuer2016}. 
Those pipelined approaches usually suffer from error propagation due to their independent models.
Thus, joint models for intent classification and slot-filling have been proposed to overcome this issue and to improve sentence-level semantics via mutual enhancement between those two tasks \cite{Weld2022}.
For those joint models, multiple approaches were explored such as conditional random fields \cite{Jeong2008}, convolutional neural networks \cite{Xu2013}, recurrent neural networks \cite{Guo2014, HakkaniTuer2016, Liu2016}, slot-gated models \cite{Goo2018}, attention mechanisms \cite{Chen2016, Liu2016}, pre-trained Transformer-based \cite{Vaswani2017} models \cite{Chen2019, Castellucci2019, Wang2020, Qin2021, Han2021} or graph convolutional network \cite{Tang2020}.

For the English language, joint models are classically evaluated on freely available benchmarks annotated with intents and concepts: ATIS \cite{Hemphill1990} and SNIPS \cite{Coucke2018}.

In the French language, less resources are available.
The ATIS dataset has been extended to French within the MultiATIS++ corpus \cite{Xu2020} by translating the manual transcriptions of the original English corpus.
However, the resulting ATIS FR dataset has no audio support available.
The MEDIA SLU dataset \cite{BonneauMaynard2005}, a native French corpus, has been actively used by the French research community and has been free for academic research since 2020. 
A study \cite{Bechet2019} showed that the MEDIA slot-filling task was one of the most challenging benchmarks among the publicly available ones.
Unfortunately, it is annotated only with concepts and not with intent.

This paper presents an updated version of the MEDIA dataset enhanced with intent annotations using a semi-automatic approach.
In addition, it presents the first results of SLU experiments on this enhanced dataset using joint models for intent classification and slot-filling.

\section{The MEDIA Benchmark}
\label{sec:mediabenchmark}

The \textit{MEDIA Evaluation Package} \citelanguageresource{MEDIA} is distributed by ELRA. 
The corpus is composed of recorded phone calls for hotel booking. It is dedicated to semantic information extraction from speech in the context of human-machine dialogues collected by using the Wizard-of-Oz method~\cite{BonneauMaynard2005}. 
The dataset represents $1258$ official recorded dialogues from $250$ different speakers and about $70$ hours of conversations. 
Only the users’ turns are annotated with both manual transcriptions and complex semantic annotations (concepts), and used in this study. 
The dataset was split into Train, Dev, and Test sets.
Each concept is represented by an attribute and detailed with a specifier. 
The semantic dictionary includes $83$ attributes and $19$ specifiers, which results in $1121$ possible attribute/specifier pairs. 
The MEDIA corpus is available in a \textit{full} or a \textit{relax scoring} version. 
In the second, attributes are simplified by excluding the specifiers. 
The number of different concepts for each set and version is presented in Table~\ref{tab:shortstats}.

Recently, \citet{Laperriere2022} proposed an updated version of the MEDIA dataset.
In addition to multiple corrections, it possesses other relevant characteristics compared to the ELRA-distributed MEDIA version.
Firstly, ELRA distributes MEDIA with two segmentation systems for audio files. 
\citet{Laperriere2022} processed MEDIA using the less commonly used segmentation system.
They also removed blank audio signals.
This segmentation system tends to create shorter utterances in augmented numbers than the most commonly used one.
The first and second lines of Table \ref{tab:shortstats} present the differences in utterances' numbers between those segmentation systems.
Secondly, they made a clear choice for managing truncated words.
The original dataset contains truncated words - words partially audible.
For example, the word "merci" (thanks in English) can be written "mer(ci)" if only the first syllable is audible on audio.
\citet{Laperriere2022} chose to keep a truncated version of those words, using the asterisk symbol '*' on the truncated part.
Our example, "mer(ci)" becomes "mer*".
Finally, \citet{Laperriere2022} noticed an available but unused, manually annotated data in the distributed MEDIA corpus.
They used it to create a second test set named Test2. 
The authors are currently working with ELRA to distribute this updated version through their catalog. While waiting for an official nomination of this version, we will cite it as \SB{} in this paper.
\begin{table*}
\begin{center}
\begin{tabularx}{2\columnwidth}{|l|X|X|X|X|}

	\hline
        \textbf{Set} & \textbf{Train} & \textbf{Dev} & \textbf{Test} & \textbf{Test2} \\
	\hline
        \textbf{Nb. utterances *} & 12916 & 1259 & 3518 &  \cellcolor{gray!20} \\
        \hline
        \textbf{Nb. utterances **} & 13712 & 1367 & 3767 &  4002 \\
	\hline
        \textbf{Concepts lexicon} (\textit{full scoring}) & 144 & 104 & 125 & 129 \\
        \hline
        \textbf{Concepts lexicon} (\textit{relax scoring}) & 73 & 63 & 71 & 71 \\
        \hline
\end{tabularx}
\caption{User's utterances characteristics of the MEDIA dataset. The number of utterances for the most commonly used segmentation system is presented in line 1 (\textbf{*}), while the version resulting from the second segmentation system, also used by \SB{} version, is presented in line 2 (\textbf{**}).}
\label{tab:shortstats}
\end{center}
\vspace*{-0.7cm}
\end{table*}

\section{Annotating The MEDIA Benchmark with Intents}

To our knowledge, the MEDIA dataset was never annotated with intents. Unlike other benchmark datasets such as ATIS \cite{Hemphill1990} or SNIPS \cite{Coucke2018}, only slots were considered.
In the context of creating a touristic dialogue system for a European project - the \HA{} project - a multilingual language understanding module capable of detecting intents and slots from utterances was envisioned.
A combination of multiple datasets, including MEDIA, was suggested for training a joint model.
But to use the MEDIA dataset for this module, we needed a version annotated with intents.
For this purpose, we defined a list of $11$ intents after carefully examining the dataset content. Some utterances can be associated with multiple intent tags, separated by the hashtag sign (\textit{\#}).
Details of this list, examples, and counter-examples will be available in an annotation guide. 
We present, as follows, how this enhanced MEDIA version containing intent annotations was obtained. Intent annotations will be available in a public repository \footnote{\url{https://github.com/Ala-Na/media\_benchmark\_intent\_annotations}}.

\subsection{Methodology}
\label{sec:methodotritraining}

Annotating a dataset can be a tremendous task. To shorten the time consumption and annotator efforts, we used a tri-training approach \cite{Zhou2005}.
Tri-training is an episodic inductive semi-supervised method \cite{Engelen2020} aiming at improving classification system performances by adding unlabeled data.
It uses a triad of classifiers trained on different training datasets. 
On each episode of the algorithm, those classifiers attribute \textit{pseudo-labels} \cite{Chen2019} to unlabeled data.
When two classifiers of the triad agree on a \textit{pseudo-label}, the corresponding \textit{pseudo-labeled} data is added to the third model's training set.
Classifiers can continue their training on the updated training sets. 
The tri-training algorithm stops when no change can be observed in the learning of all classifiers of the triad.

Recently \citet{Boulanger2022} shown that tri-training could be used in a low resource setting on a Named Entity Recognition (NER) task.
The authors used subsets of the ConLL 2003 English \cite{TjongKimSang2003} and I2B2 \cite{Uzuner2011} datasets to simulate a low-resource setting and train a triad of taggers with the tri-training algorithm \cite{Zhou2005, Ruder2018}. 
Taggers were Transformer-based BERT models \cite{Devlin2019} with a classifier architecture.
Unlabeled data was generated using sentence generation and sentence completion with a GPT-2 model~\cite{Radford2019} for a ratio of 20 synthetic data for one natural data.
The F-measure of the triad on the original test sets was evaluated.
Compared to models trained only on the subset of natural data, results were globally positive, with an average lowest gain of $0.71$ points on ConLL with $1000$ natural data and an average highest gain of $4.32$ on I2B2 with a subset of $50$ natural data.

We decided to use a similar system to train and evaluate triads of classifiers. The best triad will be kept to annotate the MEDIA dataset with intent.

\subsubsection{Datasets For Tri-training}
\label{sec:datatritraining}

A portion of manually annotated data is needed to train and evaluate triads of classifiers.
To this purpose, we used a transcribed version of the MEDIA dataset resulting from the most commonly used segmentation system, with truncated words kept as entire words ("mer(ci)" is written "merci").
For convenience, this version will be cited as MEDIA original.
A subset of randomly chosen utterances from the original training set and others picked explicitly for their content were manually annotated by one person following our intent tagging guide.
This annotation was realized out of context: each utterance was treated without considering previous ones in the dialogue.
$1551$ utterances were manually annotated for tri-training, with $1240$ constituting a train set, $124$ for a dev set, and $187$ for a test set.

Though the different intents were distributed evenly among those tri-training sets as much as possible, an imbalance effect between our classes can be observed in Table \ref{tab:distrib}.
This is likely a representation of an imbalance affecting the whole dataset.

\begin{table}[!ht]
\begin{center}
\begin{tabularx}{\columnwidth}{|l|X|X|X|X|}

    \hline
    \textbf{Set} & \textbf{train} & \textbf{dev} & \textbf{test} & \textbf{Total} \\
    \hline
    cancellation & 15 & 1 & 1 & 17\\
    \hline
    incomprehension & 6 & 1 & 4 & 11\\
    \hline
    discourse\_marker & 38 & 6 & 5 & 49\\
    \hline
    modification & 7 & 1 & 1 & 9\\
    \hline
    thanking & 47 & 5 & 6 & 58\\
    \hline
    information & 114 & 11 & 19 & 144\\
    \hline
    affirmative\_answer & 392 & 42 & 52 & 486\\
    \hline
    indecisive\_answer & 9 & 1 & 3 & 13\\
    \hline
    negative\_answer & 362 & 35 & 57 & 454\\
    \hline
    booking & 352 & 30 & 48 & 430\\
    \hline
    greeting & 43 & 8 & 6 & 57\\
    \hline

\end{tabularx}
\caption{Intent tags distribution in a subset of the MEDIA training set used for tri-training. This subset is cut into train, dev, and test set. Intents' combinations are not shown.}
\label{tab:distrib}
\end{center}
\vspace*{-0.7cm}
\end{table}

\subsubsection{Experimental Protocol}
\label{sec:protocoltritraining}
Our use case differs from \citet{Boulanger2022} work, as we have a lot of non-annotated natural data. 
We adapted their code by turning off the synthetic data generation and modifying the classifier for a \textit{multi-label} intent classifier.
It uses the final hidden state of the [CLS] special token combined with a Sigmoid layer and a threshold value of $0.5$ to determine whether the input sentence can be associated with each intent. 

We used two French Transformers \cite{Vaswani2017} models: CamemBERT \cite{Martin2020}, a model derived from RoBERTA \cite{Zhuang2021}, and FrALBERT \cite{Cattan2021}, a compact model derived from ALBERT \cite{Lan2020}. 
We evaluated two comparable versions, trained on $4$ gigabytes (GB) of text from the Wikipedia website: CamemBERT-base-Wikipedia-4GB\footnote{https://huggingface.co/camembert/camembert-base-Wikipedia-4GB} and FrALBERT-base\footnote{https://huggingface.co/qwant/fralbert-base}. 
\citet{Cattan2022} previously demonstrated that classifiers based on those models had good SLU performances on the MEDIA test dataset for the task of slot-filling.

Before the tri-training algorithm, a random sampling of $1000$ data among the $1240$ constituting our tri-training train set is made for each model of the triad. 
Fine-tuning our models on this data portion will decrease the chances that the three classifiers will output the same results. 
Though we can discuss that $1000$ among $1240$ may not offer enough variability, it reduces the event that one classifier may not be fine-tuned on intent tags poorly represented in our tri-training train set.

The algorithm is tried on a maximum of $30$ episodes, though it stops once no change is observed on a validation metric. 
Hyper-parameters are fixed with a learning rate at $1\text{e-}5$, a train batch size of $16$, and a dropout value of $0.1$. 
The number of maximal epochs per episode is $1000$, with an early stopping system of $20$ epochs of patience. 

Classifiers' performances are evaluated during training using an exact match ratio (EMR) of intents on the tri-training dev set.
The EMR is similar to accuracy but stricter as it ignores partially correct labels \cite{Sorower2010}. 
Once the tri-training algorithm stopped, EMR, precision, recall, and F-measure (or F1 score) are evaluated on the test set presented in Table~\ref{tab:distrib}. 
Those performances are calculated on the ensemble of predictions from the triad of models.

\subsubsection{Evaluation}

Results of our experiments are shown in Table~\ref{tab:restri}.
Most experiences stopped after $3$ or $4$ episodes of tri-training. 
Triads using the CamemBERT model get better results than triads using FrALBERT. They outperformed them by $7.17$ points on the EMR and $5.09$ points on the F-measure.
They also have less variability in their results, with a standard deviation oscillating between $0.33$ to $0.70$ on the different metrics against $0.81$ to $1.62$ for FrALBERT. 

\begin{table}[!ht]
\begin{center}
\begin{tabularx}{\columnwidth}{|l|X|X|X|X|}

    \hline
    \textbf{Transformer} & \textbf{EMR} & \textbf{Pre.} & \textbf{Rec.} & \textbf{F1}\\
    \hline
    CamemBERT & 92.09 & 95.29 & 93.48 & 93.73 \\
    & $\pm0.45$ & $\pm0.70$ & $\pm0.36$ & $\pm0.33$ \\
    \hline
    FrALBERT & 84.92 & 90.86 & 87.97 & 88.64 \\
    & $\pm1.62$ & $\pm0.81$ & $\pm1.44$ & $\pm1.37$\\
    \hline
\end{tabularx}
\caption{Multi-label intent classification performances with tri-training algorithm using a portion of MEDIA dataset. Results are obtained on all predictions of the triad. The mean and standard deviation error on five seeds are presented for each type of Transformer used. For precision (Pre.), recall (Rec.), and F-measure (F1), values are sample-averaged.}
\label{tab:restri}
\end{center}
\end{table}
\vspace{-0.4cm}

Following those results, we looked deeper into the performances of our best CamemBERT-based triad of models, which are presented in Table~\ref{tab:restribest}. 
This triad will be kept to annotate the MEDIA dataset with intents automatically. 
The triad obtains an EMR of $92.51$ and a sample-averaged F-measure of $93.85$.
Looking at the performance of macro F-measure, it drops to $58.99$. 
This macro F-measure seems strongly influenced by an important proportion of false negatives in some labels, with a macro recall of $60.98$.
In contrast, false positives are scarce, with a macro precision of $93.77$.
Those false negatives mainly concern tags with few examples in our test set presented in Table ~\ref{tab:distrib} as they don't affect the sample average of recall and F-measure as much.

\subsubsection{Discussion, Annotations, And Corrections}

Some changes could have been considered to improve our tri-training system. For example, using another metric than EMR to perform early stopping, such as macro or weighted F-measure, or a system considering concept labels. However, we didn't explore those possibilities as we didn't know if the performances on our test set were representative of the corrections needed to obtain a whole corpus accurately annotated with intents.

This work still represents a first approach towards using a tri-training algorithm with Transformers-based classifiers to annotate a dataset, though improvements could be applied. 

\begin{table}[!ht]
\begin{center}
\begin{tabularx}{\columnwidth}{|l|X|X|}

    \hline
    \textbf{Metric} & \textbf{Average} & \textbf{Value} \\
    \hline
    \textbf{EMR} & \cellcolor{gray!10} & 92.51 \\
    \hline
    \textbf{Precision} & sample & 95.19 \\
        & macro & 93.77 \\
    \hline
    \textbf{Recall} & sample & 93.85 \\
        & macro & 60.98 \\
    \hline
    \textbf{F1} & sample & 93.85 \\
        & macro & 58.99 \\
    \hline
    
\end{tabularx}
\caption{Multi-label intent classification performances on all predictions of our best triad of models kept to annotate the MEDIA dataset with intents. This triad uses a pre-trained CamemBERT-base-Wikipedia-4GB model.}
\label{tab:restribest}
\end{center}
\vspace*{-.3cm}
\end{table}
\begin{table}[!ht]
\begin{center}
\begin{tabularx}{\columnwidth}{|l|X|X|X|X|}

    \hline
    \textbf{Set} & \textbf{Train} & \textbf{Dev} & \textbf{Test} & \textbf{Total} \\
    \hline
    cancellation & 32 & 1 & 15 & 48\\
    \hline
    incomprehension & 273 & 30 & 94 & 397\\
    \hline
    discourse\_marker & 282 & 40 & 113 & 435\\
    \hline
    modification & 115 & 10 & 31 & 156\\
    \hline
    thanking & 713 & 100 & 200 & 1013\\
    \hline
    information & 1611 & 159 & 401 & 2171\\
    \hline
    affirmative\_answer & 4325 & 419 & 1190 & 5934\\
    \hline
    indecisive\_answer & 37 & 5 & 9 & 51\\
    \hline
    negative\_answer & 1315 & 88 & 344 & 1747\\
    \hline
    booking & 5437 & 522 & 1410 & 7369\\
    \hline
    greeting & 717 & 101 & 206 & 1024\\
    \hline
\end{tabularx}
\caption{Intent tags distribution in the preliminary version of the MEDIA dataset annotated with intents. Intents' combinations are not shown.}
\label{tab:distribfinal}
\end{center}
\vspace*{-.4cm}
\end{table}

Since our goal was to simplify annotator works, we kept the \textit{pseudo-labels} for which our best triad obtained consensus.
Sometimes, different consensus could be obtained at various episodes for the same utterance, meaning that one sentence could have more than one set of \textit{pseudo-labels}.
For those cases, one of the sets of labels was randomly chosen.
For each combination of \textit{pseudo-labels}, corresponding utterances were presented to the annotator, which had to invalidate erroneous intents.
Utterances with none or erroneous \textit{pseudo-labels} were re-annotated. 
There were $3122$ fully or partially erroneous intents ($19.51$\% of the $16005$ \textit{pseudo-labeled} data) and $137$ non-\textit{pseudo-labeled} data.

Intents tags distribution of the final version obtained is shown in Table~\ref{tab:distribfinal}. We can see that an imbalanced effect, already observed in our annotated data used for tri-training in Table~\ref{tab:distrib}, is still there. 

\subsection{Annotation Of The \SB{} Version}

The \SB{} version was also annotated.
For the Train, Dev, and Test sets, the methodology used differed from the one described in Section~\ref{sec:methodotritraining} as we already had the intents associated with each utterance.
A matching on textual content of utterances was made to retrieve intents when possible.
Manual annotations were necessary when sentence lengths differed, or truncated words were present.
Concerning truncation and to reuse our example from Section~\ref{sec:mediabenchmark}: the term "merci" ("thanks") with only the first syllable audible is textually transcribed to "mer*" in the \SB{} version.
It could correspond to the beginning of other words, like "mercredi" ("Wednesday").
Without the full word, the \textit{thanking} intent may be ignored. 

For the second test set (Test2), a similar methodology similar to Section~\ref{sec:methodotritraining} was used.
To this purpose, the previously annotated \SB{} Train, Dev, and Test sets were used as such for the tri-training algorithm.
A triad of classifiers reaching an EMR of $86.22$\% was kept.
On the $4002$ utterances composing the Test2 set, $56$ were not \textit{pseudo-labeled} by the best triad ($1.40$\% of the $4002$ data), while $289$ intents among the \textit{pseudo-labeled} were erroneous ($7.22$\% of the $3946$ \textit{pseudo-labeled} data). 

The annotation results in the enhanced \SB{} version show that the intents have an equivalent distribution between Train, Dev, Test, and Test2 datasets.
The most common intent is \textit{booking} followed by \textit{affirmative\_answer}, \textit{information} and \textit{negative\_answer}.
While the least common intent is \textit{cancellation} followed by \textit{indecisive\_answer}.

\section{Experiments On Manual Transcriptions}
\label{sec:expmanualtranscript}

Using the enhanced MEDIA dataset, we present a baseline by training and evaluating models on manual transcriptions, performing the joint training of intent classification and slot-filling tasks.

\subsection{Neural Architecture}
\label{JointArch}
The BERT model architecture for joint intent classification and slot-filling \cite{Chen2019} is a modified version of a BERT model \cite{Devlin2019}. 
It uses Softmax activation functions to determine the intent and slots of each utterance.
For intent classification, the final hidden state of the [CLS] token is fed to a Softmax layer.
For slot-filling, the final state of each first sub-token of a word is provided to a Softmax layer to determine which concept can be associated with the word. 
The model is fine-tuned by optimizing the sum of cross-entropy losses for both tasks. 

We modified this architecture to perform a multi-label intent classification instead of a multi-class classification and keep the slot-filling part, using a Sigmoid layer and a threshold value of $0.5$.
The probability ${P}_{i}$ for an input to be associated with an intent $i$ passing $h_{[CLS]}$, the Transformer's last hidden state for [CLS] token, to a layer of weight $W^{i}$ and bias $b^{i}$ is defined as:
$P_{i} = sigmoid(W^{i}h_{[CLS]} + b^{i}) > 0.5$. 
A binary cross-entropy loss replaces the cross-entropy loss previously used for intent classification.
The model is fined-tuned on the sum of the binary and non-binary cross-entropy losses for intent classification and slot-filling, respectively.

\subsection{Experimental Protocol}
\label{sec:protocolnlu}

For the slot-filling task, we used a BIO-tagging format.
Performances are evaluated in terms of micro F-measure, commonly used for joint models \cite{Weld2022}, and Concept Error Rate (CER), the official metric used in the MEDIA campaign \cite{BonneauMaynard2006}.
For later comparison with experiences on Automatic Speech Recognition (ASR) outputs, we follow the micro F-measure calculated on multi-hot vectors of concepts present in expected and predicted annotations.

For intent classification, when there are multiple intents, we concatenate them using pound marks~(\#). In most joint models, this task's performance is evaluated using the accuracy \cite{Weld2022}. 
As we use a multi-label classification system, the accuracy as proposed by \citet{Godbole2004} and EMR were evaluated.

The sentence-level semantic frame accuracy (SFA) - corresponding to the number of utterances with perfectly predicted intent and slots divided by the number of sentences - commonly used for joint models \cite{Weld2022}, is also evaluated.

We replaced the BERT model with French models. 
We choose CamemBERT base trained on $135$ GB of text from CCNET (CamemBERT-base-CCNET\footnote{https://huggingface.co/camembert/camembert-base-ccnet}) \cite{Martin2020} as well as the previously used CamemBERT-base-Wikipedia-4GB and FrALBERT-base. 
Those models demonstrated state-of-the-art, or close to, results on slot-filling using MEDIA manual transcriptions \cite{Ghannay2020, Cattan2022}.
We also choose a French BERT model, FlauBERT, fine-tuned for a few epochs on ASR data (FlauBERT-oral-ft\footnote{https://huggingface.co/nherve/flaubert-oral-ft}) \cite{Herve2022} which demonstrated close to state-of-the-art performances on MEDIA ASR outputs \cite{Pelloin2022}.

Following \citet{Cattan2022} study, we used a population-based training (PBT) algorithm \cite{Jaderberg2017} to explore hyperparameters. 
We considered a number of training epochs between $5$ and $100$, a batch size in the interval of $8$ and $32$, and a learning rate ranging between $1$ and $5\text{e-}5$. 
To select the best trial among a population with PBT, the algorithm uses the mean value of slot-filling F-measure summed with intent classification accuracy.
For the \SB{} version, we evaluated our performances only on the first Test dataset.

\subsection{Results on manual transcriptions}
\label{sec:resnlutab}

Performances on original relax, \SB{} relax, and \SB{} full versions of the MEDIA dataset are displayed in Table~\ref{tab:resnlu}.

On the original relax version, CamemBERT-base-Wikipedia-4GB obtains the best results on intent classification with an accuracy of $93.98$ and an intent EMR of $91.84$.
On slot-filling, CamemBERT-base-CCNET get the best results with an F-measure of $88.52$ and a CER of $8.68$, reaching a SFA of $76.26$. 
On the \SB{} relax version, CamemBERT-base-CCNET performs the best on intent EMR with $89.78$. 
But for intent accuracy and slot-filling, FlauBERT-oral-ft obtains the best results with $92.10$ of intent accuracy, $87.75$ of slot-filling F-measure, $9.18$ of CER, reaching an SFA of $73.29$.
On the \SB{} full version, FlauBERT-oral-ft obtains the best results on intent classification with an accuracy of $92.31$ and an EMR of $89.97$.
CamemBERT-base-CCNET intent classification performances are close, and its slot-filling performances are the highest, with an F-measure of $85.33$ and a CER of $11.61$. 
But for the SFA, CamemBERT-base-Wikipedia-4GB performs slightly better than the rest with $72.15$.

Looking at multi-hot concept vectors F-measure, results are logically better than slot-filling F-measure.
This metric is, however, only present for later comparison with experiments using ASR outputs.

Concerning concept annotation versions, we can logically observe that models perform better on the relax version than the full version for the task of slot-filling.
For example, there is a difference of $2.42$ points of F-measure between the best results obtained on \SB{} relax and full version, in favor of the relax version.
More surprisingly, all models perform better on both tasks with the original relax than on the \SB{} relax version.
This could be explained by truncated words in the latter, making it more difficult to recognize semantic concepts and intents.

Comparing our work to previous studies, we cannot reach the best CER result obtained on the original MEDIA relax version by \citet{Ghannay2020} with a value of $7.56$ for CamemBERT-base-CCNET.
We are also behind \citet{Cattan2022} results with F-measures of $89.9$, $90.0$, and $89.8$ as well as CERs of $7.5$, $8.4$ and $8.6$ for CamemBERT-base-CCNET, CamemBERT-base-Wikipedia-4GB and FrALBERT-base respectively.
Though our architectures - and training conditions for \citet{Cattan2022} - are close, this can be explained by using a training objective considering intent classification and slot-filling in this work, making our training less focused on the slot-filling task.

\begin{table*}
\begin{center}
\begin{tabularx}{2\columnwidth}{lXXXXXX}

    \hline
    & \multicolumn{2}{>{\hsize=\dimexpr2\hsize+4\tabcolsep+2\arrayrulewidth\relax}X}{\quad\textbf{Intent}} & \multicolumn{3}{>{\hsize=\dimexpr3\hsize+6\tabcolsep+3\arrayrulewidth\relax}X}{\quad\quad\quad\textbf{Slot-filling}} & \\
    \textbf{Model} & \textbf{Acc.} & \textbf{EMR} & \textbf{F1} & \textbf{F1mh} & \textbf{CER} & \textbf{SFA} \\
    \hline
    \hline
    \multicolumn{6}{c}{\textbf{MEDIA original, relax}} \\
    \hline
    \hline
    CamemBERT-base-CCNET & 93.87 & 91.79 & 88.52 & \textbf{95.97} & 8.68 & \textbf{76.26}\\
    \citet{Ghannay2020} & \cellcolor{gray!10} & \cellcolor{gray!10} & 89.37 & \cellcolor{gray!10} & \textbf{7.56}  & \cellcolor{gray!10} \\
    \citet{Cattan2022} & \cellcolor{gray!10} & \cellcolor{gray!10} & 89.9 & \cellcolor{gray!10} & 7.5 & \cellcolor{gray!10} \\
    \hline
    CamemBERT-base-Wikipedia-4GB & \textbf{93.98} & \textbf{91.84} & 87.93 & 95.41 & 9.34 & 75.58\\
    \citet{Cattan2022} & \cellcolor{gray!10} & \cellcolor{gray!10}  & \textbf{90.0} & \cellcolor{gray!10} & 8.4 & \cellcolor{gray!10} \\
    \hline
    FlauBERT-oral-ft & 93.66 & 91.19 & 87.93 & 95.63 & 8.95 & 76.04 \\
    \hline
    FrALBERT-base & 92.27 & 89.88 & 84.24 & 93.66 & 13.14 & 72.12\\
    \citet{Cattan2022} & \cellcolor{gray!10} & \cellcolor{gray!10} & 89.8 & \cellcolor{gray!10} & 8.4 & \cellcolor{gray!10} \\
    \hline
    \hline
    \multicolumn{6}{c}{\textbf{\SB{}, relax}} \\
    \hline
    \hline
    CamemBERT-base-CCNET & 91.87 & \textbf{89.78} & 86.95 & 94.66 & 10.33 & 72.68 \\
    CamemBERT-base-Wikipedia-4GB & 91.25 & 88.66 & 86.88 & 94.88 & 10.24 & 72.60\\
    FlauBERT-oral-ft & \textbf{92.10} & 89.73 & \textbf{87.75} & \textbf{95.41} & \textbf{9.18} & \textbf{73.29} \\
    FrALBERT-base & 90.71 & 88.37 & 82.48 & 92.94 & 14.71 & 69.18\\  
    \hline
    \hline
    \multicolumn{6}{c}{\textbf{\SB{}, full}} \\
    \hline
    \hline
    CamemBERT-base-CCNET & 92.28 & 89.73 & \textbf{85.33} & \textbf{92.87} & \textbf{11.61} & 72.13\\
    CamemBERT-base-Wikipedia-4GB & 91.81 & 89.25 & 85.24 & 92.42 & 12.11 & \textbf{72.15}\\
    FlauBERT-oral-ft & \textbf{92.31} & \textbf{89.97} & 84.26 & 92.34 & 12.68 & 71.54 \\
    FrALBERT-base & 90.64 & 88.29 & 80.10 & 90.38 & 17.40 & 68.04\\
    \hline
    \hline
\end{tabularx}
\caption{Best model performances with population-based training on different versions of the MEDIA test dataset manual transcriptions. Performances are evaluated with accuracy (Acc.) and EMR for intent classification. For slot-filling, performances are given in terms of F-measure (F1), F-measure on concepts multi-hot vectors (F1mh) and CER. The SFA is also evaluated.}
\label{tab:resnlu}
\vspace{-0,7cm}
\end{center}
\end{table*}

\section{SLU Experiments}
Using the enhanced MEDIA dataset, we present baselines of SLU performances for cascade and end-to-end systems, performing the joint training of intent classification and slot-filling tasks.
We evaluate both approaches' performances using the previous metrics followed in section \ref{sec:protocolnlu}, except for slot-filling F-measure.

\subsection{Cascade}

The cascaded approach consists of using two components to solve specific problems separately.
First, an ASR system maps speech signals to automatic transcriptions.
This is then passed on to the joint model presented in section~\ref{JointArch}, which predicts semantic information (slots and intents) from the automatic transcriptions.
The ASR model used for the cascade approach is made of the LeBenchmark FR 3k large \cite{Evain2021} speech encoder, followed by $3$ bi-LSTM layers plus one linear layer of $1024$ neurons.
Both speech encoder and Bi-LSTM layers are updated with an Adam optimizer of $0.0001$ learning-rate, while the linear output layer uses an Adadelta optimizer of $1.0$ learning-rate. 
The Connectionist Temporal Classification (CTC) greedy loss function is optimized for $100$ epochs, aiming for the best Word Error Rate (WER) possible, as we obtained $9.49$\% of WER on \SB{} and $10.51$\% of WER on MEDIA original dataset with our system.

Cascade system results are shown in Table~\ref{tab:resslucascade}.
On the original relax version, results tendencies follow the ones in Section \ref{sec:resnlutab} with CamemBERT-base-Wikipedia-4GB reaching the best intent accuracy ($92.43$) and CamemBERT-base-CCNET reaching the best multi-hot concept vectors F-measure ($93.82$).
On \SB{} relax version, FlauBERT-oral-ft reaches the best results for intent accuracy with $90.40$, slot-filling multi-hot F-measure with $93.40$, CER with $11.93$ and SFA with $64.96$.
On intent EMR, CamemBERT-base-CCNET performs slightly better with $88.00$.
On \SB{} full version, CamemBERT-base-CCNET performs better than other models with an intent accuracy of $90.86$, an intent EMR of $88.29$, a slot-filling multi-hot F-measure of $91.06$ and a CER of $14.18$. It reaches $64.14$ of SFA.
Though FlauBERT-oral-ft was specifically fine-tuned on ASR outputs, this doesn't confer an absolute advantage in this cascade system.


\begin{table*}
\begin{center}
\begin{tabularx}{2\columnwidth}{lXXXXX}

    \hline
    & \multicolumn{2}{>{\hsize=\dimexpr2\hsize+4\tabcolsep+2\arrayrulewidth\relax}X}{\quad\qquad\textbf{Intent}} & \multicolumn{2}{>{\hsize=\dimexpr2\hsize+4\tabcolsep+2\arrayrulewidth\relax}X}{\quad\textbf{Slot-filling}} & \\
    \textbf{Model} & \textbf{Accuracy} & \textbf{EMR} & \textbf{F1mh} & \textbf{CER} & \textbf{SFA} \\
    \hline
    \hline
    \multicolumn{6}{c}{\textbf{MEDIA original, relax}} \\
    \hline
    \hline
    CamemBERT-base-CCNET & 92.07 & 89.82 & \textbf{93.82} & 13.93 & \textbf{65.69} \\
    \citet{Ghannay2021} & \cellcolor{gray!10} & \cellcolor{gray!10} & \cellcolor{gray!10} & \textbf{11.2} & \cellcolor{gray!10} \\
    \hline
    CamemBERT-base-Wikipedia-4GB & \textbf{92.43} & \textbf{90.08} & 93.29 & 15.03 & 65.49 \\
    \hline
    FlauBERT-oral-ft & 92.28 & 89.77 & 93.12 & 14.19 & 65.55 \\
    \citet{Pelloin2022} & \cellcolor{gray!10} & \cellcolor{gray!10} & \cellcolor{gray!10} & 11.98$\pm0.68$ & \cellcolor{gray!10} \\
    \hline
    FrALBERT-base & 90.81 & 88.18 & 91.14 & 19.97 & 62.91 \\
    \hline
    \hline
    \multicolumn{6}{c}{\textbf{\SB{}, relax}} \\
    \hline
    \hline
    CamemBERT-base-CCNET & 90.16 & \textbf{88.00} & 92.74 & 12.78 & 64.43 \\
    CamemBERT-base-Wikipedia-4GB & 89.39 & 86.75 & 92.90 & 13.19 & 64.00 \\
    FlauBERT-oral-ft & \textbf{90.40} & 87.95 & \textbf{93.40} & \textbf{11.93} & \textbf{64.96} \\
    FrALBERT-base & 89.29 & 86.86 & 91.00 & 17.81 & 62.01 \\  
    \hline
    \hline
    \multicolumn{6}{c}{\textbf{\SB{}, full}} \\
    \hline
    \hline
    CamemBERT-base-CCNET  & \textbf{90.86} & \textbf{88.29} & \textbf{91.06} & \textbf{14.18} & \textbf{64.14} \\
    CamemBERT-base-Wikipedia-4GB & 90.62 & 88.13 & 90.60 & 15.19 & 63.66 \\
    FlauBERT-oral-ft & 90.23 & 87.76 & 90.68 & 15.11 & 63.10 \\
    FrALBERT-base & 88.89 & 86.25 & 88.22 & 20.16 & 61.24 \\
    \hline
    \hline
    
\end{tabularx}
\caption{Cascade results for MEDIA datasets using a LeBenchmark FR 3k large speech encoder and different Transformer-based models with joint architecture. Performances are evaluated with accuracy and EMR for intent classification. For slot-filling, performances are given in terms of F-measure on multi-hot concept vectors (F1-micro) and CER. The SFA is also evaluated.}
\label{tab:resslucascade}
\vspace{-0.3cm}
\end{center}
\end{table*}

\subsection{End-to-end}

\begin{table*}[!h]
    \begin{center}
        \begin{tabularx}{2\columnwidth}{lXXXXX}
        
            \hline
            & \multicolumn{2}{>{\hsize=\dimexpr2\hsize+4\tabcolsep+2\arrayrulewidth\relax}X}{\quad\qquad\textbf{Intent}} & \multicolumn{2}{>{\hsize=\dimexpr2\hsize+4\tabcolsep+2\arrayrulewidth\relax}X}{\quad\textbf{Slot-filling}} & \\
            \textbf{Model} & \textbf{Accuracy} & \textbf{EMR} & \textbf{F1mh} & \textbf{CER} & \textbf{SFA} \\
            \hline
            \hline
            
            \multicolumn{6}{c}{\textbf{MEDIA original, relax}} \\
            \hline
            \hline
            SAMU-XLSR & \textbf{92.02} & \textbf{90.14} & 91.68 & 16.01 & 71.57 \\
            SAMU-XLSR $_{IT \oplus FR}$ & 91.74 & 90.02 & \textbf{92.35} & 15.44 & \textbf{72.51} \\
            LeBenchmark FR 3k large & 91.75 & 89.91 & 91.98 & 15.51 & 71.12 \\
            \citet{Pelloin2021} & \cellcolor{gray!10} & \cellcolor{gray!10} & \cellcolor{gray!10} & \textbf{13.6} & \cellcolor{gray!10} \\
            \hline
            \hline
            
            \multicolumn{6}{c}{\textbf{\SB{}, relax}} \\
            \hline
            \hline
            SAMU-XLSR & \textbf{90.74} & \textbf{88.85} & 90.01 & 15.28 & 68.50 \\ 
            SAMU-XLSR $_{IT \oplus FR}$ & 90.53 & 88.64 & 90.65 & 15.16 & 70.83 \\ 
            LeBenchmark FR 3k large & 90.18 & 87.98 & \textbf{90.77} & \textbf{15.08} & \textbf{71.12} \\ 
            \hline
            \hline
            
            \multicolumn{6}{c}{\textbf{\SB{}, full}} \\
            \hline
            \hline
            SAMU-XLSR & 90.52 & 88.69 & 88.88 & 18.50 & 69.72 \\ 
            SAMU-XLSR $_{IT \oplus FR}$ & \textbf{90.98} & \textbf{88.93} & \textbf{89.14} & \textbf{18.30} & \textbf{70.63} \\
            LeBenchmark FR 3k large & 90.07 & 88.02 & 87.99 & 19.67 & 69.06 \\
            \citet{Laperriere2023} & \cellcolor{gray!10} & \cellcolor{gray!10} & \cellcolor{gray!10} & 18.5 & \cellcolor{gray!10} \\
            \hline
            \hline
            
        \end{tabularx}
        \caption{End-to-end results for MEDIA datasets with different speech encoders (SAMU-XLSR, SAMU-XLSR$_{IT \oplus FR}$, LeBenchmark FR 3K large) in terms of accuracy and EMR for intent classification, and F-measure on multi-hot concept vectors (F1mh) and CER for slot-filling. The SFA is also evaluated.}
        \label{tab:rese2e}
        \vspace{-0.6cm}
    \end{center}
\end{table*}

The end-to-end system aims to develop a single system directly optimized to extract semantic information from speech without using intermediate speech transcriptions.
Our end-to-end model consists of a fine-tuned speech encoder (original SAMU-XLSR \cite{Khurana2022}, specialized SAMU-XLSR$_{IT \oplus FR}$ \cite{Laperriere2023} leading to the best results for \SB, or LeBenchmark FR 3k large), followed by two different decoding blocks of $3$ bi-LSTM layers of $1024$ neurons, here for segments contextualization.
Each is followed by a fully connected layer of the same dimension, activated with LeakyReLU and a Softmax function. 
One branch is optimized to yield the intents of the audio segments, while the other performs the original slot-filling task of MEDIA.
We optimized our CTC greedy loss functions on $100$ epochs with the same optimizers used in the cascade approach, apart from the linear layer of the intent classification optimizer, which has its learning rate set to $0.1$. 
The sum of both losses is defined as $loss = \frac{1}{4}*loss(intent) + loss(slot)$.
We load checkpoints from our best CER with no significant intent accuracy deterioration. 

Before settling with this architecture, we considered using a single decoding block for both tasks. 
Experiments showed that the CER significantly worsened when its decoding weights were also being updated for the intent task. 
We also tried out different learning rates for our optimizers and different loss ratios and metric-based checkpoints.

Table \ref{tab:rese2e} gives the results on intent classification and slot-filling tasks with this end-to-end architecture and different speech encoders.
Considering only our end-to-end models, we obtained the state-of-the-art results on \SB{} full slot-filling task with $18.30$\% of CER, besides the margin with \citet{Laperriere2023} $18.5$\% CER not being significant, a difference of 0.8 points of CER being necessary for it to be relevant. 
To our knowledge, CERs for the MEDIA task are degraded when using an end-to-end architecture compared to a cascade one. 
However, this is not the case for intent classification, as shown by the best accuracy scores and EMRs obtained with those systems for all datasets, an exception remaining in the $0.4$\% improvement of MEDIA original's best accuracy score with a cascade approach.
The gap between Table \ref{tab:rese2e} and Table \ref{tab:resslucascade} intent classification results might, however, not be impactful enough, considering the results obtained for the joint slot-filling task, leading to globally better cascade results. 
At last, we can affirm a better joint optimization on both tasks with a largely better SFA score for each MEDIA version with our end-to-end models. 


\section{Conclusion}

In this paper, we present an enhanced version of the MEDIA benchmark dataset with intent annotations.
We expect to broaden the use of this French dataset for more SLU tasks.
We also present the first experimental results on this enhanced dataset using joint models for intent classification and slot-filling. 

We presented different baseline systems for joint intent classification and slot-filling applied, whether on manual transcriptions, automatic transcriptions (cascade), or speech signals (end-to-end).
Experimental results on manual and automatic transcriptions could not reach the previous state-of-the-art results for the task of slot-filling but are still competitive.
End-to-end models performing joint optimization seem to obtain better scores on both tasks than cascade models.

\section{Acknowledgements}
This paper was partially funded by the European Commission through the HumanE-AI-Net project under grant number 952026 and by the \textit{Multiligual SLU for Contextual Question Answering (MuSCQA)} project from "France Relance" of French government funded by French National Research Agency (ANR), grant number: ANR-21-PRRD-0001-01.
This work was performed using HPC resources from GENCI–IDRIS (Grant 2023-A0131013834).

\section{Limitations} 

This study has potential limitations.
As intent annotations' were mainly made by only one annotator, no inter-annotator agreement (IAA) could be calculated.
Other annotators should inspect the current version of intent annotations to reach an IAA. 
The annotator was a female engineer working at LISN whose first language was French.

\section{Bibliographical References}
\label{reference}

\bibliographystyle{lrec-coling2024-natbib}
\bibliography{lrec}

\section{Language Resource References}
\label{lr:ref}
\bibliographystylelanguageresource{lrec-coling2024-natbib}
\bibliographylanguageresource{languageresource}

\appendix

\begin{figure*}[!ht]
\begin{center} 
    \scalebox{0.75}{
    \begin{tikzpicture}[framed, rounded corners]
        \centering
        \node (words) {\textbf{Utterance}};
        \node (w1) [right=0.7cm of words] {book};
        \node (w2) [right=1.5cm of w1] {one};
        \node (w3) [right=2.5cm of w2] {room};
        \node (w4) [right=1.5cm of w3] {in};        \node (w5) [right=0.3cm of w4] {Marseille};
        \node (w6) [right=0.3cm of w5] {for};
        \node (w7) [right=1.5cm of w6] {two};
        \node (w8) [right=2.5cm of w7] {people};
        \node (slots) [below=0.3cm of words] {\textbf{Slot labels}};
        \node (bio) [below=0cm of slots] {(BIO-tagging format)};
        \node (s1) [below=0.3cm of w1] {B-task};
        \node (s2) [below=0.3cm of w2] {B-number\_of\_rooms};
        \node (s3) [below=0.3cm of w3] {I-number\_of\_rooms};
        \node (s4) [below=0.3cm of w4] {O};
        \node (s5) [below=0.3cm of w5] {B-city};
        \node (s6) [below=0.3cm of w6] {O};
        \node (s7) [below=0.3cm of w7] {B-number\_of\_guests};
        \node (s8) [below=0.3cm of w8] {I-number\_of\_guests};
        \node (int) [below=0.3cm of bio] {\textbf{Intent label}};
        \node (intent) [right=1cm of int] {booking};
        \draw [->] (w1) -- (s1);
        \draw [->] (w2) -- (s2);
        \draw [->] (w3) -- (s3);
        \draw [->] (w4) -- (s4);
        \draw [->] (w5) -- (s5);
        \draw [->] (w6) -- (s6);
        \draw [->] (w7) -- (s7);
        \draw [->] (w8) -- (s8);
    \end{tikzpicture}
    }
\end{center}
\caption{Simplistic SLU example. Each word of the utterance is associated with a slot label, while the utterance as a whole is associated with an intent label.}
\label{fig:slu}
\end{figure*}

\section{Appendix A: Example Of SLU Notation}

A simplistic example of SLU notation, illustrated with intent and slot labels, is presented in Figure~\ref{fig:slu}.
\newline

\section{Appendix B: Annotation Manual Extract (In French)}

\textit{
The following text presents an extract of the first draft of the annotation manual used to annotate the tri-training subset and correct the pseudo-labels.
The extract contains the list of proposed intentions accompanied by an explanation, examples, and counter-examples.
It is written in French, the language of the MEDIA benchmark.
\newline
}

L'annotation a été réalisée hors-contexte, en ne disposant que des tours de paroles des utilisateurs.\\

\textbf{Liste des intentions proposées :}\\

Onze différentes étiquettes d'intentions ont été distinguées.
Ces intentions ont pour vocation d'identifier le but de la requête de l'utilisateur.
Pour un système de dialogue, l'identification de cette intention permettrait de formuler une réponse adaptée.
En compréhension du dialogue, les intentions se veulent complémentaires aux concepts.

Pour un exemple type \textit{"je souhaite réserver une chambre à Marseille pour deux personnes"}, l'intention pouvant être associée est la réservation d'une chambre d'hôtel.
Les concepts identifieront plutôt les paramètres de la réservation (lieu, nombre de chambres, nombre de personnes, etc.).
Pour un système de dialogue destiné à la réservation d'hôtel confronté à cet exemple, la finalité serait d'effectuer une réservation correspondant aux attentes de l'utilisateur grâce à cette compréhension des intentions et des concepts.

Pour chaque intention considérée, une brève description est proposée et complétée par des exemples et contre-exemples.

Les noms des intentions proposées sont :
\begin{itemize}
    \setlength\itemsep{0em}
    \item reponse\_affirmative
    \item reponse\_negative
    \item reponse\_indecise
    \item saluer
    \item remercier
    \item marqueur\_discursif
    \item incomprehension
    \item reservation
    \item renseignements
    \item annulation
    \item modification
\end{itemize}


\subsection{reponse\_affirmative}

L'intention \textit{reponse\_affirmative} concerne les énoncés où la réponse apportée par l'utilisateur est une réponse affirmative ou une confirmation.
Dans le contexte d'une conversation téléphonique avec un serveur vocal, ce type de réponse peut être fréquemment rencontrée  suite à une question fermée.

\subsubsection{Exemples}

Dans la plupart des cas, il peut s'agir d'une réponse courte consistant en un unique marqueur d'accord :

\begin{itemize}
    \setlength\itemsep{0em}
    \item \textit{oui}
    \item \textit{voilà}
    \item \textit{parfait}
    \item \textit{ok}
    \item \textit{exactement}
    \item \textit{entendu}\\
\end{itemize}

Dans d'autres cas, la réponse peut être plus longue et comprendre des détails contextuels ou comprendre des marqueurs discursifs.
Elle peut aussi ne pas comprendre de marqueur d'accord bien qu'une confirmation soit présente :

\begin{itemize}
    \setlength\itemsep{0em}
    \item \textit{sans problème}
    \item \textit{oui et c' est c' est de bon standing oui}
    \item \textit{ben oui j' ai pas j' ai pas le choix oui}
    \item \textit{ben très bien je prends}
    \item \textit{vas-y effectue}
    \item \textit{oui s' il vous plaît ça sera vachement bien j' en ai vraiment vraiment envie et c' est important}
    \item \textit{le jour de l' arrivée d'accord}
\end{itemize}

\subsubsection{Contre-exemples}

Des contre-exemples accompagnés d'explications sont disponibles en Table~\ref{tab:ce_rep_aff}.

\begin{table*}[ht!]
\begin{center}
\resizebox{\textwidth}{!}{
\small
\begin{tabularx}{2\columnwidth}{|X|X|X|}
	\hline
        \textbf{Énoncé} & \textbf{Intention} & \textbf{Justification}\\
	    \hline
        \textit{alors je prends l' hôtel Saint-Charles à soixante quinze euros avec le parking} & reservation & Selon le contexte, l'expression "alors je prends" pourrait correspondre à une réponse affirmative. Cependant la suite de la phrase correspond a des critères de réservation. Il est possible que l'utilisateur annonce sa décision parmi plusieurs options qui lui ont été proposées.\\
        \hline
        \textit{et ben je vais quand même prendre celle euh avec le chien donc la plus chère} & reservation & Selon le contexte, le début de l'énoncé peut correspondre à une réponse affirmative. Néanmoins, la suite laisse suggérer que l'utilisateur énonce son choix parmi plusieurs options.\\
        \hline
        \textit{bon alors l' hôtel Ibis réservons l' hôtel Ibis du trois au cinq juin} & reservation & La phrase pourrait correspondre a une réponse affirmative. Néanmoins, sans contexte, on peut imaginer que l'utilisateur annonce sa décision entre plusieurs choix. Des critères de réservation sont aussi énoncés.\\
        \hline
        \textit{non c' est ok} & reponse\_negative & "c'est ok" (marqueur d'accord) pourrait correspondre à une réponse affirmative mais vient ici appuyer la réponse négative précédente.\\
        \hline
        \textit{d' accord et il y a le téléphone} & renseignements & Le terme "d' accord" semble ici utilisé comme un marqueur discursif, car il est suivi d'une demande de renseignements.\\
        \hline
\end{tabularx}
}
\caption{Contre-exemples pour l'intention \textit{reponse\_affirmative}}
\label{tab:ce_rep_aff}
\end{center}
\end{table*}


\subsection{reponse\_negative}

L'intention \textit{reponse\_negative} regroupe les énoncés où l'utilisateur fournit une réponse négative ou qu'il signale son refus.

Comme pour l'intention \textit{reponse\_affirmative}, ce type de réponse peut être fréquemment rencontrée dans une conversation téléphonique, notamment lorsqu'elles suivent une question fermée.

\subsubsection{Exemples}

Il peut s'agir d'une réponse courte consistant en un unique marqueur de désaccord.
Du contexte ou des marqueurs discursifs peuvent y être ajoutés.
Un refus peut aussi être exprimé sans présence d'un marqueur de désaccord :

\begin{itemize}
    \setlength\itemsep{0em}
    \item \textit{non}
    \item \textit{c' est sans importance}
    \item \textit{euh j' ai pas d' exigence d' exigence particulière}
    \item \textit{euh non}
    \item \textit{euh non ça ira j' ai pris note}
    \item \textit{non c' est bon c' est tout}
    \item \textit{non c' est parfait}
\end{itemize}

\subsubsection{Contre-exemples}

Un contre-exemple accompagné de son explication est disponible en Table~\ref{tab:ce_rep_neg}.

\begin{table*}[hbt!]
\begin{center}
\resizebox{\textwidth}{!}{
\small
\begin{tabularx}{2\columnwidth}{|X|X|X|}
	\hline
        \textbf{Énoncé} & \textbf{Intention} & \textbf{Justification}\\
	    \hline
        \textit{aucun} & reservation & Le mot "aucun" pourrait éventuellement servir de réponse négative à une question. Il est cependant plus probable que cette réponse apporte des détails à une réservation.\\
        \hline
	\end{tabularx}
}
\caption{Contre-exemple pour l'intention \textit{reponse\_negative}}
\label{tab:ce_rep_neg}
\end{center}
\end{table*}


\subsection{reponse\_indecise}

L'intention \textit{reponse\_indecise} est la 3ème catégorie de réponses rencontrée et concerne les énoncés où l'utilisateur exprime principalement son indécision.

Si une telle réponse est présentée à un système de dialogue, celui-ci pourrait demander à l'utilisateur de clarifier sa décision.

\subsubsection{Exemples}

Une indécision peut être exprimée sous plusieurs formes. L'utilisateur peut l'exprimer clairement ou laisser la décision au serveur vocal :

\begin{itemize}
    \setlength\itemsep{0em}
    \item \textit{hum celle que vous voulez}
    \item \textit{n je sais pas}
    \item \textit{euh je ne sais pas}
\end{itemize}

\subsubsection{Contre-exemples}

Aucun contre-exemple n'est disponible pour cette catégorie d'intention.


\subsection{saluer}

Comme dans la plupart des systèmes de dialogues, une catégorie d'intention \textit{saluer} est crée. Cette catégorie vise à détecter les salutations de l'utilisateur.
Il peut s'agir d'une salutation servant à initier la conversation, ou au contraire à la terminer.

\subsubsection{Exemples}

Les énoncés correspondant à cette intention correspondent aux différentes formules de salutations existantes :

\begin{itemize}
    \setlength\itemsep{0em}
    \item \textit{bonjour}
    \item \textit{à tout à l' heure}
    \item \textit{au revoir madame}
    \item \textit{de rien au revoir}
\end{itemize}

\subsubsection{Contre-exemples}

Aucun contre-exemple n'est disponible pour cette catégorie d'intention.


\subsection{remercier}

L'intention \textit{remercier}, rencontrée dans de nombreux systèmes de dialogue, vise à identifier un remerciement de la part de l'utilisateur.
La détection de cette intention invite à retourner une formule de courtoisie.

\subsubsection{Exemples}

Les exemples correspondant à cette catégorie d'intention comprennent les formules de remerciement communément utilisées dans la langue française :

\begin{itemize}
    \setlength\itemsep{0em}
    \item \textit{je vous remercie beaucoup}
    \item \textit{merci}
    \item \textit{euh je vous remercie bien}
\end{itemize}

\subsubsection{Contre-exemples}

Un contre-exemple accompagné de son explication est disponible en Table~\ref{tab:ce_rem}.
\newpage

\begin{table*}[hbt!]
\begin{center}
\resizebox{\textwidth}{!}{
\small
\begin{tabularx}{2\columnwidth}{|X|X|X|}
	\hline
        \textbf{Énoncé} & \textbf{Intention} & \textbf{Justification}\\
	    \hline
        \textit{non merci} & reponse\_negative & Le mot "merci" suivant le marqueur de désaccord "non" et n'appuyant pas d'autres arguments par la suite est une formule de politesse fréquente en langue française, auquel il ne convient pas forcément de retourner une autre formule de politesse.\\
        \hline
\end{tabularx}
}
\caption{Contre-exemple pour l'intention \textit{remercier}}
\label{tab:ce_rem}
\end{center}
\end{table*}


\subsection{marqueur\_discursif}

Sous l'intention \textit{marqueur\_discursif} sont rangés les énoncés n'apportant ni information, ni demande, ni réponse de la part de l'utilisateur.
Ils n'invitent pas à une réponse adaptée de la part du serveur.

Les données ayant été récoltées sous forme d'enregistrements audio, de nombreuses énoncés sont concernées et consistent parfois en un ou plusieurs marqueurs discursif, d'où le nom de cette catégorie.

Cette intention ne peut être combinée à une autre.

\subsubsection{Exemples}

Il peut s'agir d'interjection ou de formule de politesse n'invitant pas à une réponse particulière ("pas de quoi", "je vous en prie").
Des énoncés courts et ne comportant pas d'informations utiles, mais entrant dans la catégorie des marqueurs discursifs usuellement rencontrés dans la langue française, peuvent aussi concernés.
Pour un système de dialogue, il conviendrait d'attendre un nouvel énoncé de la part de l'utilisateur ou de notifier l'utilisateur de sa présence (avec un réponse telle que "oui ?") :

\begin{itemize}
    \setlength\itemsep{0em}
    \item \textit{euh}
    \item \textit{hein}
    \item \textit{ah}
    \item \textit{hum hum}
    \item \textit{ah excusez-moi}
    \item \textit{je vous en prie}
    \item \textit{hum pas de quoi}
    \item \textit{je peux une minute}
    \item \textit{ben écoutez}
    \item \textit{c' est noté}
    \item \textit{alors}
\end{itemize}

\subsubsection{Contre-exemples}

Des contre-exemples accompagnés d'explications sont disponibles en Table~\ref{tab:ce_md}.

\begin{table*}[hbt!]
\begin{center}
\resizebox{\textwidth}{!}{
\small
\begin{tabularx}{2\columnwidth}{|X|X|X|}
	\hline
        \textbf{Énoncé} & \textbf{Intention} & \textbf{Justification}\\
	    \hline
        \textit{euh oui j' aimerais changer de dates} & modification & Une interjection est présente. Cependant, la lecture de l'énoncé laisse suggérer que l'utilisateur veut modifier sa réservation. Un but est présent derrière cette phrase, il ne peut donc s'agir de l'intention \textit{marqueur\_discursif}.\\
        \hline
        \textit{et je voudrais} & incomprehension & Ici, un énoncé court et incomplet est présent. Il ne s'agit pas d'un marqueur discursif usuellement rencontré dans la langue française. L'énoncé aurait besoin d'être complété pour que le système puisse en extraire une information utile. Une demande de reformulation ou d'apport de précision à l'utilisateur serait nécessaire.\\
        \hline
\end{tabularx}
}
\caption{Contre-exemples pour l'intention \textit{marqueur\_discursif}}
\label{tab:ce_md}
\end{center}
\end{table*}


\subsection{incomprehension}

Sous l'étiquette \textit{incomprehension} sont rangées les énoncés où un problème de communication est présent.
Ce problème peut survenir chez l'interlocuteur (qui ne comprend pas le serveur ou qui ne formule pas quelque chose d'exploitable) ou chez le serveur (qui ne répond pas à l'attente de l'utilisateur ou rencontre des problèmes techniques).
Dans le second cas, nous ne disposons que d'une observation indirecte puisque seuls les tours de paroles de l'utilisateur sont disponibles.

Le corpus textuel ayant été récupéré par transcription d'enregistrements téléphoniques, certaines énoncés concernées peuvent contenir l'interjection "allô" utilisée parfois lorsque la qualité sonore d'un appel diminue ou que des problèmes de communication sont rencontrés.
Cette interjection étant utilisée dans d'autres contextes, elle ne saurait constituer à elle seule un critère d'appartenance à cette catégorie d'intention.
"allô" peut aussi être utilisé en début de conversation téléphonique.

Les problèmes de communication pouvant frustrer l'utilisateur, celui-ci peut exprimer son mécontentement.
L'utilisateur peut aussi chercher à clarifier la situation.

Parfois, c'est l'utilisateur qui peut formuler une énoncé dont l'intention est incompréhensible ou hors-sujet, même avec prise en compte du contexte.
Ces formulations peuvent aussi comprendre des morceaux de phrases incomplètes, sans information utile ou exploitable par le système, mais qui ne rentre pas dans la catégorie des marqueurs discursifs usuellement rencontrés en langue française.

\subsubsection{Exemples}

\begin{itemize}
    \setlength\itemsep{0em}
    \item \textit{n' importe quoi}
    \item \textit{je comprends pas moi}
    \item \textit{à cette vous vous bégayez ou euh}
    \item \textit{dans la chambre en plus je je veux pas que les gens}
    \item \textit{excusez-moi allô}
    \item \textit{ça colle pas ça}
    \item \textit{viennent piétiner dans ma chambre ça vous le comprenez c' est pas possible donc alors c' est là euh vous faites ça euh c' est pas possible hein je vais téléphoner je vais écrire à votre maison moi allô allô} 
    \item \textit{tais toi chéri}
    \item \textit{de problème}
\end{itemize}

\subsubsection{Contre-exemples}

Des contre-exemples accompagnés d'explications sont disponibles en Table~\ref{tab:ce_inc}.

\begin{table*}[hbt!]
\begin{center}
\resizebox{\textwidth}{!}{
\small
\begin{tabularx}{2\columnwidth}{|X|X|X|}
        \hline
        \textbf{Énoncé} & \textbf{Intention} & \textbf{Justification}\\
	    \hline
        \textit{hôtel comment} & renseignements & L'utilisateur semble ne pas avoir compris le nom de l'hôtel mais cherche ici à l'obtenir, il s'agit donc plutôt d'une demande de renseignements.\\
        \hline
        \textit{alors je voudrais euh pour les trois premiers jours de juillet à Paris près de Saint-Germain-Des-Prés euh une chambre pour un couple euh moins de cent vingt euros la nuit et avec un accès handicapés euh à l' hôtel quoi un accès handicapés handicapés} & reservation & La fin de la phrase avec la partie "euh à l' hôtel quoi" pourrait laisser suggérer une incompréhension de la part de l'utilisateur. La lecture de la phrase dans son intégralité montre cependant qu'il clarifie le sens de sa demande.\\
        \hline
        \textit{dans le dernier hôtel Ibis euh je sais pas quoi} & reservation & Le "je ne sais pas quoi" peut laisser penser à une incompréhension de la part de l'utilisateur mais cette expression sert ici à désigner l'hôtel souhaité.\\
        \hline
        \textit{dans le deuxième hôtel euh Stanislas je sais plus combien à quarante euros} & reservation & Le "je ne sais plus" est utilisé pour designer l'hôtel choisi par l'utilisateur et complémentaire d'autres informations ("deuxième hôtel", "Stanislas", "à quarante euros").\\
        \hline
\end{tabularx}
}
\caption{Contre-exemples pour l'intention \textit{incomprehension}}
\label{tab:ce_inc}
\end{center}
\end{table*}


\subsection{reservation}

L'intention \textit{reservation} concerne tous ce qui prête à la réservation, c'est-à-dire aux demandes de réservation ou à l'apport de critères de réservation.
Les critères de réservation sont variés et peuvent concerner (de manière non exhaustive) : Le nom de l'hôtel, la localisation de l'hôtel, la période de réservation souhaitée, le nombre de chambre, le nombre de personnes, le nombre de nuits, la fourchette de prix souhaitée, la présence d'un service au sein de l'hôtel, l'accessibilité de l'hôtel, l'équipement désiré dans la chambre, la réservation de chambres voisines, le nombre d'étoiles de l'hôtel, les services à proximité de l'hôtel, etc.
Ces critères de réservation sont détectés par la tache d'identification des concepts (ou \textit{slot filling}).

Les énoncés concernés peuvent n'inclure qu'une demande de réservation, pour initier la recherche répondant aux besoins de l'interlocuteur.

Dans le contexte d'un système de dialogue où le serveur pose des questions à l'utilisateur pour obtenir des précisions, certains énoncés peuvent être brefs et ne contenir qu'un critère de réservation ou qu'une information exploitable dans son context.
Par exemple, "deux" peut être la réponse à une question rapportant au nombre de chambres, au nombre de personnes, au nombre de nuits, etc.
Les énoncés peuvent aussi concerner la rectification ou modification implicite d'un critère de réservation par rapport à ce que l'utilisateur avait précisé auparavant ou par rapport à ce qui lui est proposé.
Ils peuvent aussi être formulés de manière interrogative.
Par exemple, les phrases "c'est trop cher" ou "vous n'avez pas quelque chose de moins cher ?", suite à une proposition de chambre dans un hôtel dont le prix a été énoncé, impliquent que l'utilisateur souhaiterait se voir proposer une réservation similaire mais à un tarif moins élevé.

Parfois, les énoncés concernés peuvent aussi être l'annonce d'un choix parmi des options proposées par le serveur vocal
L'utilisateur peut ainsi spécifier quel hôtel et quels critères de réservation ont sa préférence.

\subsubsection{Exemples}

\begin{itemize}
    \setlength\itemsep{0em}
    \item \textit{euh réserver euh dans un hôtel} 
    \item \textit{réserver un hôtel}
    \item \textit{réservation}
    \item \textit{je souhaite réserver à Paris place Gambetta les trois premiers jours d' octobre une chambre simple à moins de cinquante euros}
    \item \textit{je souhaite deux chambres c' est-à-dire deux couples dont un avec enfant}
    \item \textit{alors j' aurais voulu une autre chambre euh pas trop chère avec euh aussi chambre euh ensoleillée}
    \item \textit{à Arles}
    \item \textit{Ibis}
    \item \textit{un enfant}
    \item \textit{une}
    \item \textit{cent neuf}
    \item \textit{en campagne}
    \item \textit{un accès handicapés}
    \item \textit{les quatre derniers euh quatre}
    \item \textit{vue mer}
    \item \textit{mois d' août}
    \item \textit{c' est c' est trop cher}
    \item \textit{proche d' une salle de cinéma}
    \item \textit{près du lac}
    \item \textit{alors je prends l' hôtel Saint-Charles à soixante quinze euros avec le parking}
    \item \textit{euh ben écoutez je crois que je vais prendre le le premier hôtel l' hôtel de du désir à quarante cinq euros}
\end{itemize}

\subsubsection{Contre-exemples}

Des contre-exemples accompagnés d'explications sont disponibles en Table~\ref{tab:ce_res}.

\begin{table*}[hbt!]
\begin{center}
\resizebox{\textwidth}{!}{
\small
\begin{tabularx}{2\columnwidth}{|X|X|X|}
	\hline
        \textbf{Énoncé} & \textbf{Intention} & \textbf{Justification}\\
	    \hline
        \textit{le prix le nombre d' étoiles} & renseignements & Ici, il ne s'agit pas de critères de sélection de l'hôtel car aucune valeur n'est attribuée. Cet énoncé peut être perçu comme une demande de renseignements. \\
        \hline
        \textit{et total} & renseignements & En absence de contexte, il semble s'agir plutôt d'une demande de renseignements dont la nature n'est pas précisée.\\
        \hline
        \textit{euh non si peut-être le le le la piscine la plus proche de l' hôtel il y en a pas dedans mais il y a sûrement une à côté} & renseignements \& reponse\_indecise & Il s'agit d'une demande de renseignement sur un service de proximité et non d'un critère de réservation. Elle suit une réponse indécise "non si peut-être".\\
        \hline
\end{tabularx}
}
\caption{Contre-exemples pour l'intention \textit{reservation}}
\label{tab:ce_res}
\end{center}
\end{table*}


\subsection{renseignements}

Cette intention concerne les énoncés où l'utilisateur exprime vouloir des informations, ou formule une demande de renseignements sur un à plusieurs hôtels.
Ces renseignements peuvent concerner l'adresse d'un hôtel, son accessibilité, ses services, son nombre d'étoiles, ses modes de règlement, les services de proximité, etc.

Contrairement à l'intention \textit{reservation}, on attend que le serveur réponde à une question au lieu d'inclure un nouveau critère de réservation.

Parfois, une demande de renseignements peut être formulée sans que la nature des renseignements souhaités soit présente dans la phrase.
De la même manière, un énoncé peut contenir uniquement la nature de ces renseignements et la demande est alors implicite.

\subsubsection{Exemples}

\begin{itemize}
    \setlength\itemsep{0em}
    \item \textit{plus de détails pour l' hôtel Campanile}
    \item \textit{plus des détails}
    \item \textit{obtenir d' autres informations}
    \item \textit{à l' hôtel Passy pouvez-vous répéter le tarif}
    \item \textit{le prix}
    \item \textit{je voudrais savoir s' il y a un accès handicapés et s' il y a une baignoire et le prix de ces chambres}
    \item \textit{je voudrais savoir s' il y a un tennis dans un de ces deux hôtels}
    \item \textit{et comment j' aurais la conf}
    \item \textit{on peut régler en carte bleue}
    \item \textit{est-ce que je dois vous envoyer un acompte}
    \item \textit{si euh éventuellement l' hôtel accepte des animaux}
    \item \textit{hum est-ce qu' il y a le téléphone}
    \item \textit{y a-t-il des animations le soir}
\end{itemize}

\subsubsection{Contre-exemples}

Des contre-exemples accompagnés d'explications sont disponibles en Table~\ref{tab:ce_rens}.

\begin{table*}[hbt!]
\begin{center}
\resizebox{\textwidth}{!}{
\small
\begin{tabularx}{2\columnwidth}{|X|X|X|}
	\hline
        \textbf{Énoncé} & \textbf{Intention} & \textbf{Justification}\\
	    \hline
        \textit{ah excusez-moi alors je voudrais un hôtel à Paris à côté de la tour Eiffel pour les d pour les nuits du sept au huit juin trois chambres individuelles avec un sauna dans l' hôtel et euh une chambre à deux cents francs deux cents euros maxi et avec un chien un gros chien est-ce que c' est possible} & reservation & Une question est posée en fin d'énoncé, mais ce sont bien des critères de réservation qui ont été annoncés. La question appelle à chercher une réservation correspondant à ces critères.\\
        \hline
        \textit{euh vous n' avez rien de moins cher avec accès handicapés vous n' avez rien de moins cher} & reservation & Ici, l'énoncé peut être perçu comme l'évolution implicite d'un critère de réservation : L'utilisateur souhaite un prix moins élevé que celui actuellement proposé par le serveur. De plus, un autre critère de réservation (présence d'un accès handicapé) est aussi demandé.\\
        \hline
\end{tabularx}
}
\caption{Contre-exemples pour l'intention \textit{renseignements}}
\label{tab:ce_rens}
\end{center}
\end{table*}


\subsection{modification}

L'étiquette \textit{modification} comprend les énoncés où une modification dans une réservation est souhaitée.
La modification peut concerner la réservation effectuée précedemment ou la réservation en cours.

Elle peut parfois se recouper avec l'intention \textit{reservation}, lorsque l'utilisateur modifie les critères d'une réservation.
On estime cependant qu'un énoncé peut correspondre à l'intention \textit{modification} dès qu'une demande explicite de modification est réalisée.

Des termes tels que "autre", "modifier" ou "changer" peuvent être utilisés.

\subsubsection{Exemples}

\begin{itemize}
    \setlength\itemsep{0em}
    \item \textit{euh je voudrais changer un critère} 
    \item \textit{je désire changer réserver dans un autre hôtel}
    \item \textit{je veux modifier les dates vingt trois septembre au vingt neuf septembre}
    \item \textit{changer de dates}
    \item \textit{et y en a pas d' autres}
    \item \textit{hum ben euh ouais une autre date}
\end{itemize}

\subsubsection{Contre-exemples}

Des contre-exemples accompagnés d'explications sont disponibles en Table~\ref{tab:ce_mod}.

\begin{table*}[hbt!]
\begin{center}
\resizebox{\textwidth}{!}{
\small
\begin{tabularx}{2\columnwidth}{|X|X|X|}
	\hline
        \textbf{Énoncé} & \textbf{Intention} & \textbf{Justification}\\
	    \hline
        \textit{non à Lorient} & reponse\_negative \& reservation & Avec l'utilisation d'une réponse négative, il s'agit plutôt d'une rectification.\\
        \hline
        \textit{non non dans l' hôtel Ti Moana c' est ça allô} & reponse\_negative \& reservation & Avec l'utilisation d'une réponse négative, il s'agit plutôt d'une rectification. Des problèmes de communications semblent présents mais sont secondaires.\\
        \hline
        \textit{euh réserver euh deux chambres dans un autre hôtel} & reservation & Ici, la mention d'un "autre hôtel" peut concerner une seconde réservation.\\
        \hline
        \textit{d'accord et l' autre} & renseignements & Le terme "autre" est utilisé précédé d'un article défini. "autre" désigne donc un objet énoncé préalablement dans la conversation.\\
        \hline
\end{tabularx}
}
\caption{Contre-exemples pour l'intention \textit{modification}}
\label{tab:ce_mod}
\end{center}
\end{table*}


\subsection{annulation}

L'intention \textit{annulation} comprends les demandes ou ordres d'annulation explicites.
Il peut s'agir d'annuler une réservation en cours ou réalisée préalablement à l'appel actuel.

\subsubsection{Exemples}

\begin{itemize}
    \setlength\itemsep{0em}
    \item \textit{alors à ce moment-là j' annule tout parce que je n' ai je ne peux pas réserver pour quelque chose que je ne connais pas à ce moment-là vous annulez tout le numéro cent trente quatre cent quatre vingt douze} 
    \item \textit{annulez tout le numéro cent trente quatre cent quatre vingt douze}
    \item \textit{bien bon ben alors euh on annule tout}
    \item \textit{vous annulez}
    \item \textit{donc annulation}
    \item \textit{bien ben écoutez je regrette euh je j' annule j' annule ma demande}
    \item \textit{ok bon ben je réserve pas de chambre alors}
    \item \textit{euh oui et ben ça serait tout je réserve pas}
\end{itemize}

\subsubsection{Contre-exemples}

Aucun contre-exemple n'est disponible pour cette catégorie d'intention.


\subsection{Combinaisons d'intentions}

A l'exception de l'intention \textit{marqueur\_discursif}, qui correspond aux énoncés ou aucune autre intention n'est présente, les différentes intentions peuvent être combinées.

Dans le cadre d'un système de dialogue soumis à ces énoncés, chaque intention pourrait nécessiter une prise en charger particulière.
Par exemple, la phrase "A bientôt et merci" serait identifiée par la combinaison d'intentions \textit{saluer} ("au revoir") et \textit{remercier} ("merci").
La prise en compte de ces deux intentions pourrait permettre la formulation d'une réponse telle que "De rien et au revoir".

Plusieurs exemples de ces combinaisons sont fournis ci-dessous, sans constituer une liste exhaustive.

\subsubsection{Exemples}

Des exemples sont fournis en Table~\ref{tab:ex_comb}

\begin{table*}[hbt!]
\begin{center}
\resizebox{\textwidth}{!}{
\small
\begin{tabularx}{2\columnwidth}{|X|X|X|}
        \hline
        \textbf{Énoncé} & \textbf{Intention} & \textbf{Justification}\\
        \hline
        \textit{je ne comprends pas comment vous pouvez me donner un une réservation d' hôtel si je n' ai pas l' adresse si vous n' avez pas l' information j' annule tout} & annulation, incomprehension \& renseignements & Le "je ne comprends pas" traduit une forte incompréhension de la part de l'utilisateur. Celui-ci souhaite aussi obtenir une adresse d'hôtel, donc un renseignement. De plus, une intention d'annulation, bien que conditionnelle, est annoncée.\\
        \hline
        \textit{merci beaucoup au revoir} & remercier \& saluer & L'utilisateur remercie et salue le serveur.\\
        \hline
        \textit{donnez-moi le tarif puisque je voudrais que ça coûte moins de cinquante euros mais qu' il y ait quand même un parking privé } & renseignements \& reservation & Ici, des critères de réservation sont présents ("moins de cinquante euros", "parking privé"). Un renseignement est aussi demandé ("le tarif").\\
        \hline
        \textit{non à Marseille} & reponse\_negative \& reservation & Un marqueur de désaccord est présent et suivi d'un critère de réservation (ville).\\
        \hline
        \textit{parfait je veux réserver deux chambres individuelles chambres voisines si possible} & reponse\_affirmative \& reservation & Ici, l'utilisateur approuve le choix qui lui est présenté et ajoute des critères de réservation supplémentaires.\\  
        \hline
        \textit{non j' aimerais avoir des informations de suite pour les deux nuits le cinq et le sept à Chartres août} & renseignements, reponse\_negative \& reservation & L'utilisateur commence sa phrase par un marqueur de désaccord. Une demande de renseignements est présente, bien que la nature des renseignements ne soit pas précisée. Des critères de réservation (dates du séjour) sont aussi présentés.\\
        \hline
\end{tabularx}
}
\caption{Exemples de combinaisons d'intentions.}
\label{tab:ex_comb}
\end{center}
\end{table*}

\section{Appendix C: Annotation Of The MEDIA 2022
Version}

Details of the \SB{} version intent tags distribution are presented in Table~\ref{tab:distribmedia2022}.

\begin{table*}[hbt!]
\begin{center}
\resizebox{\textwidth}{!}{
\small
\begin{tabularx}{2\columnwidth}{|l|X|X|X|X|X|}

    \hline
    \textbf{Set} & \textbf{Train} & \textbf{Dev} & \textbf{Test} & \textbf{Test2} & \textbf{Total} \\
    \hline
    cancellation & 33 & 1 & 17 & 12 & 63 \\
    \hline
    incomprehension & 409 & 49 & 128 & 58 & 644 \\
    \hline
    discourse\_marker & 386 & 48 & 148 & 97 & 679 \\
    \hline
    modification & 127 & 11 & 33 & 23 & 194 \\
    \hline
    thanking & 691 & 99 & 199 & 175 & 1164 \\
    \hline
    information & 1704 & 158 & 408 & 524 & 2794 \\
    \hline
    affirmative\_answer & 4347 & 427 & 1209 & 1207 & 7190 \\
    \hline
    indecisive\_answer & 48 & 9 & 10 & 6 & 73 \\
    \hline
    negative\_answer & 1345 & 90 & 343 & 412 & 2190 \\
    \hline
    booking & 5777 & 575 & 1528 & 1810 & 9690 \\
    \hline
    greeting & 714 & 101 & 210 & 199 & 1224 \\
    \hline
    
\end{tabularx}
}
\caption{Intent tags distribution in the preliminary version of the \SB{} dataset \citep{Laperriere2022} annotated with intents. Intents' combinations are not shown. There is no distinction between the \textit{full} and \textit{relax} scoring versions, as they only differ on slots annotations.}
\label{tab:distribmedia2022}
\end{center}
\end{table*}

\end{document}